\documentclass[runningheads,a4paper]{llncs}

\usepackage{amssymb}
\usepackage{graphicx}

\begin{document}

\mainmatter  

\title{Combinatorial Convolutional Neural Networks for Words}

\titlerunning{Combinatorial Convolutional Neural Networks for Words}

%
%
\author{Karen Sargsyan}
\authorrunning{}

\institute{Iinstitute of Biomedical Sciences, Academia Sinica,\\
128 Sec. 2 Academia Rd., Taipei, Taiwan}

%
%

\maketitle

\begin{abstract}
The paper discusses the limitations of deep learning models in identifying and utilizing features that remain invariant under a bijective transformation on the data entries, which we refer to as combinatorial patterns. We argue that the identification of such patterns may be important for certain applications and suggest providing neural networks with information that fully describes the combinatorial patterns of input entries and allows the network to determine what is relevant for prediction. To demonstrate the feasibility of this approach, we present a combinatorial convolutional neural network for word classification.

\end{abstract}

\section{Introduction}

The standard description of how deep learning works involves feeding raw data into the input layer of a neural network and using backpropagation to update the weights of the network such that it can learn to extract meaningful features from the data. This contrasts with 'traditional' machine learning methods, which often rely on domain experts to manually design relevant features for the model to effectively learn and make predictions\cite{deepl}. As an example, one can train a convolutional neural network to classify pictures of cats and dogs by feeding it with raw pixel data from the images. In such task, network achieves \textgreater90\% accuracy on a validation dataset\cite{cats}. 

\begin{figure}
\centering
\includegraphics[height=6.2cm]{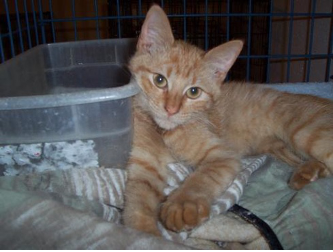}
\includegraphics[height=6.2cm]{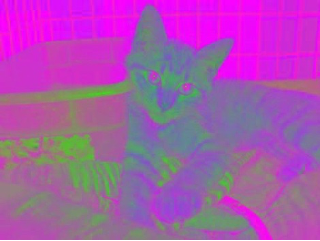}
\caption{An example of the same one-to-one pixel color transformation applied to a photo of a cat, where the human eye clearly recognizes the cat, while the deep learning model is confused.}
\label{fig:example}
\end{figure}

An interesting question is to what degree a deep learning model, when trained for a specific task, is able to identify and utilize features that remain invariant under a bijective transformation on the data entries. As an example, we may change the color for each pixel of a cat picture defined by a triplet of natural numbers (R, G, B) into some other value (R1, G1, B1) using the same fixed bijection  on every pixel, which here defines one of the possible transformations of a cat picture.  It so happens that in our cats/dogs example, the model's confidence in classifying the new image as a cat drops significantly, while a human observer would have no difficulty in doing so (Fig \ref{fig:example}). In this case, the neural network prioritizes features that include particular combinations and arrangement of colored pixels and to a lesser extent combinatorial patterns, which we here understand as features that are invariant under the bijective transformation of an input. As another example, we trained a convolutional neural network to classify papers into four categories (such as 'Sport') based on their text content, which was input to the network as raw character data\cite{zhang}. Our network achieved \textgreater88\% accuracy on a validation dataset. Next, we transformed the text in the validation dataset by replacing each letter with its corresponding letter from a randomly generated alphabet permutation, which was the same for all texts in the dataset. The accuracy dropped to 24\%. 

	The two examples given above demonstrate that, in general, neural networks do not select features based on combinatorial patterns. One possible reason for this is that the data used to train neural networks is often asymmetric and may not include all possible transformations. For instance, a transformation where each letter in a text is encoded with a possibly different letter may result in gibberish that naturally does not occur in readable texts. In the cat/dog example, the dataset is biased towards natural colors and scenes commonly found in photographs. 
 
   While deep learning in general does not provide learning based on combinatorial patterns and leans towards encoding particular trends that are over-represented in training data, it still enjoys enormous success in applications. Therefore, one may ask why to care about prediction based on combinatorial patterns? There are several possible applications of prediction based on combinatorial patterns that motivate its utility. First, some prediction tasks may involve processes that are heavily reliant on combinatorial patterns. For example, it has been shown that viruses and their hosts have similar nucleotide patterns in their genomes, beyond encoding similar amino acids. This similarity indicates that genetic sequences have similar optimization for host ribosomes, which translate them to proteins. This is a crucial enough factor to use for predicting potential new hosts for viruses or estimating their potential danger to the human population\cite{baba,iuchi}. As an additional example, the distribution of a word in the text may provide clues about whether that word is a keyword of the text\cite{key}. Similarly, the distribution of nucleotide words in RNA may signal that the nucleotide word composes an important 3D motif of RNA structure\cite{RNA}.  
   
   In certain cases, we may not know to what extent combinatorial patterns are relevant to our prediction task beforehand. Having two models - one that has learned to predict based on combinatorial patterns and another that has not - can help us determine the degree to which these patterns contribute to our ability to make predictions and understand the processes we are studying. Second, unexpected changes in input type for some applications, such as object detection in photographs, may require retraining a neural network with new data. This process may be time-consuming and unacceptable for mission-critical applications that are time-constrained. However, if a successful model has been trained using combinatorial patterns, it is inherently more robust and may require less retraining.  Additionally, one can potentially combine two types of learning in a single model to overcome the limitations of a skewed training dataset. For instance, the processes at play when following a cat using a video camera may degrade, resulting in different video quality (Fig.\ref{fig:example}). This could hinder the system's ability to react promptly to different cat situations. It is easy to imagine similar scenarios involving military drones instead.
 
	Accepting the utility of prediction based on combinatorial patterns, one may propose to extend the dataset by transforming its entries in all possible ways. However, the number of possible transformations for just our example of a cat picture makes the task of covering all or the majority of variants computationally and space-prohibitive. As an approach, one may consider using deep learning to be a guide, where we feed all the available data into the model and expect it to select appropriate features. We suggest feeding the neural network information that fully describes the combinatorial patterns of the input entries, stripping away all other information, and letting the network decide what is important for predicting. Successfully tackling the combinatorial aspects of the input objects is a prerequisite for this task and in this work, we will concentrate on words as inputs and on combinatorial patterns for words. In the next chapter we will introduce relevant mathematical concepts borrowed from \cite{morita,morita2,morita3} that let us fully describe combinatorial patterns. Then, we present examples of neural networks learning to classify words as palindrome or non-palindrome and estimate the strengths of passwords using combinatorial patterns as input. We compare their performance with a model trained on raw character-based data. Our main goal is to demonstrate the feasibility of this approach and highlight possible applications of mathematical results from the field of abstract algebra.

\section{Words and Combinatorics}

Let us first introduce some notation from \cite{morita}. Let $\alpha$ be a word over some alphabet, and let $\Omega(\alpha)$ denote the set of distinct letters appearing in $\alpha$. We define $S(\alpha)$ as the set of all subwords(i.e. contiguous substrings) of $\alpha$. By adding an extra subword/letter $\varepsilon$ to $S(\alpha)$ we construct a new set $W(\alpha) = \{\varepsilon \}\cup S(\alpha)$. One may consider $\varepsilon$ to be an empty subword of $\alpha$.

For any two subwords of $\alpha$, $\lambda = Y_{1}Y_{2}...Y_{s}$ and $\mu = Z_{1}Z_{2}...Z_{t}$, we make the matrix $(m_{ij})$, where $m_{ij} = Y_{i}$ if $Y_{i} = Z_{j}$ and $m_{ij} = \varepsilon$ otherwise.  Using this matrix we construct the associated graph, whose vertices are $(i, j)$ for all $1\leqslant i\leqslant s$ and $1\leqslant j\leqslant t$ and edges $(i, j) \to (k, l)$ if $k = i + 1, l = j + 1, m_{ij} \neq \varepsilon, m_{kl} \neq \varepsilon$. For every connected component of the graph, one may produce the associated subword of $\alpha$ using values of $(m_{ij})$ and a path on the graph starting from the vertex with the lowest index values for a given connected component. Therefore, each subword of $\alpha$, including an empty subword $\varepsilon$ might be potentially produced by a connected component of the graph. Different connected components may produce different or the same subwords. It is natural to ask how often a given subword $\nu$ is produced by connected components of the graph generated by some other subwords $\lambda$ and $\mu$. We call such counting $M(\alpha)_{\nu}(\lambda,\mu)$ the combinatorics for $\alpha$ following \cite{morita}, where the reader may find a more detailed description. With an extra set of rules:
\begin{equation}
  M(\alpha)_{\nu}(\lambda, \varepsilon) = \delta_{\nu,\varepsilon}*s,
  M(\alpha)_{\nu}(\varepsilon, \mu) = \delta_{\nu,\varepsilon}*t,
  M(\alpha)_{\nu}(\varepsilon, \varepsilon) = \delta_{\nu,\varepsilon},
\end{equation}
we obtain a map:
\begin{equation}
  M(\alpha): W(\alpha)\times W(\alpha)\times W(\alpha) \to \mathrm{Z}_{\geqslant0},\\
 (\lambda,\mu,\nu) \longmapsto M(\alpha)_{\nu}(\lambda, \mu).
\end{equation}
Our goal is to be able to tell whether two words $a$ and $b$ demonstrate the same patterns. By our definition there is a bijection $\varphi$ from $\Omega(a)$ to $\Omega(b)$ such that $a=X_{1}X_{2}...X_{r}$ and $b=\varphi(X_{1})\varphi(X_{2})...\varphi(X_{r})$. Based on theorems proven \cite{morita}, we may
no longer search for a bijection, but compare combinatorics for words of interest, as
$M(\alpha)_{\nu}(\lambda, \mu)$ combinatorics contain exactly information we are
after.

\section{Combinatorical Convolutional Neural Networks for Words}
Combinatorics, as defined in the previous chapter, allows us to provide comprehensive information on patterns in words to neural networks as raw input. Training a neural network on combinatorics for a given word may enable it to select appropriate combinatorial features for a given task of classification. One popular neural network architecture that has shown great success in text and image classification is Convolutional Neural Networks (CNNs). In CNN, for the case of an image, a special convolutional layer presents data as a three-dimensional tensor (a 3D 'matrix'), where two dimensions (number of elements along the axis) correspond to the positions of each pixel, and the third dimension corresponds to so-called filters, representing the (R, G, B) colors of the input picture. In the case of $200\times200$ pixel colored image, we get $200\times200\times3$ tensor. Convolutional layers following the input layer tend to decrease the first two dimensions and increase the number of filters. The filters of a CNN are designed to extract different features from the input data, such as edges, curves, and textures in the case of image processing. By stacking multiple convolutional layers with different filters, the network can learn hierarchical representations of the input data, where lower layers extract simple features, and higher layers capture more complex and abstract features. With combinatorics, we may utilize this approach and use $M(\alpha)_{\nu}(\lambda, \mu)$ as filters over dimensions $\lambda \times \mu$ in the input 3D tensor for a given word $\alpha$. 
One particular property, in this case, is that number of elements along each direction of the input tensor is the same and is defined by a number of subwords of the word. 

\section{Case of Word Classification into Palindromes/Non-Palindromes}
To demonstrate the feasibility of such an approach, we present a toy example of a classification task where only combinatorial patterns matter. Our goal is to classify 20-letter words into palindromes and non-palindromes. We generated 1000 palindromes and 1000 non-palindromes for the training dataset, as well as 500 examples of each class for both the validation and testing sets. Non-palindromes were randomly generated from an alphabet and checked to conform to the definition. Palindromes were generated from randomly generated words by 'palindromising' them \footnote{full code and datasets to verify the paper's results will be deposited here \url{https://github.com/karsar/wordsCCNN}.}.
\begin{figure}
\centering
\includegraphics[height=6.2cm]{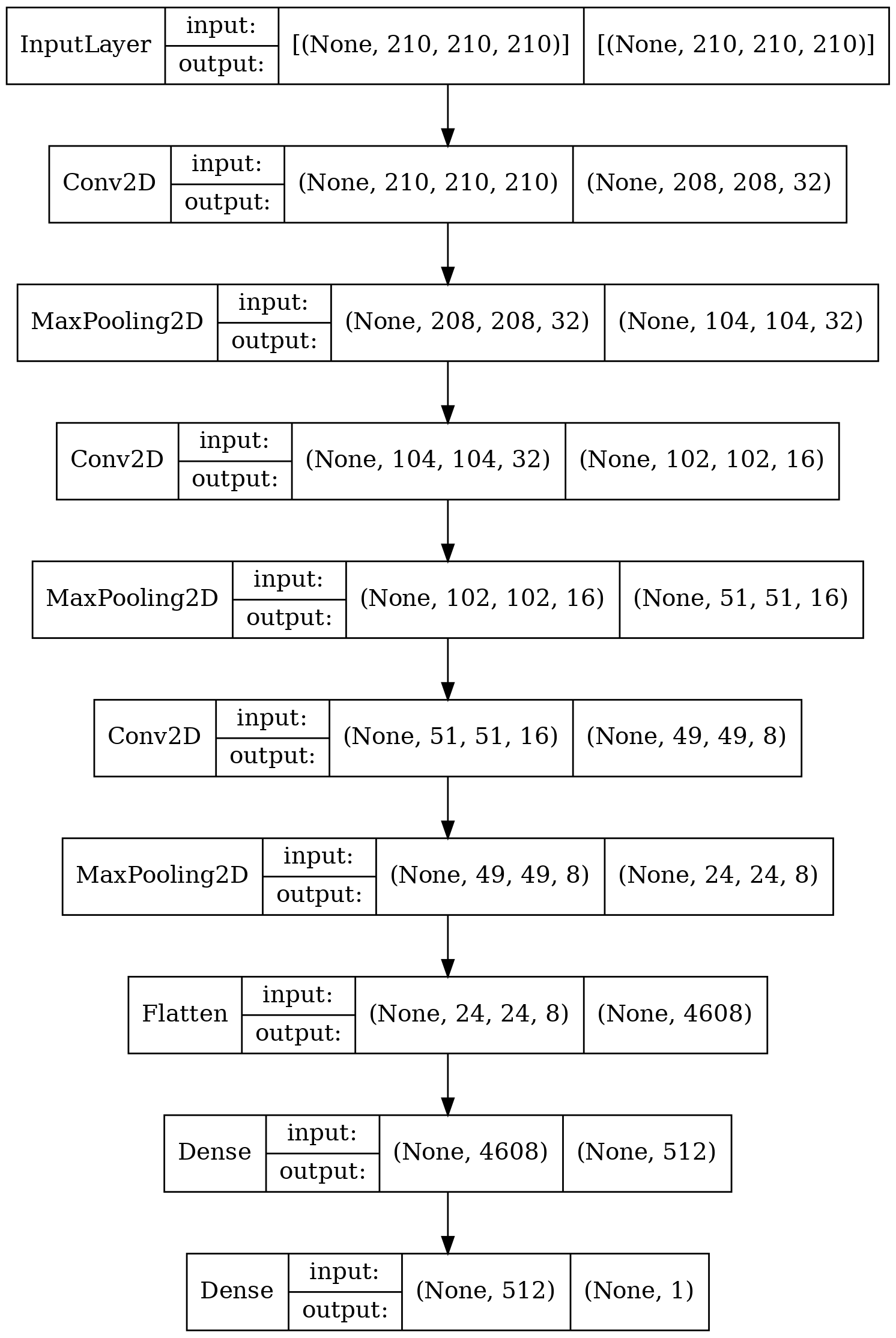}
\caption{A neural network layout for dealing 20 letter word classification task.}
\label{fig:example1}
\end{figure}

Our combinatorial CNN to classify 20-letter words into two classes is presented in Fig.\ref{fig:example1}. It consists of convolutional layers interspersed with max-pooling layers, which is the standard first choice to try. The layout is common for CNNs, with one notable difference of decreasing the number of filters as we move from layer to layer, while usually, the number grows. We trained the model using batches of 32 words and each training epoch contained 30 batches. After the first epoch accuracy on training and validation sets was $92.6\%$ and $96\%$ correspondingly. The final accuracy after 48 epochs was $100\%$ for the training and $99.17\%$ for the validation datasets. Considering both how well our first attempt to build combinatorial convolutional neural networks has gone and our goal to demonstrate feasibility, we did not pursue further possible improvements and optimizations of the network.  

As the next step, we used the previously mentioned convolutional neural network that we used to classify papers in order and applied it to learn classifying 20-letter words as palindromes/non-palindromes\cite {zhang}. It is not surprising that it achieved similar performance, for this particular task, as only combinatorial patterns were relevant. The conventional CNN was forced to learn only combinatorial patterns, which is not typically the case in other applications. Additionally, we didn't modify a neural network that was designed to classify longer texts, therefore, it has more layers and a greater capacity to learn than the one we used for the combinatorial case. 

\section{Case of Password Classification into Strong and Weak}
Another example of a classification task where only combinatorial patterns matter is password classification into weak and strong. To test whether our approach of using combinatorial CNNs works, we have randomly generated a dataset of 2000 passwords of length 15 for the training dataset and 1000 passwords of the same length for each of the validation and test datasets. In each dataset half of the passwords are weak and another half is strong. We define a password to be strong if its complexity, taking value from [0,1], calculated by python package \textit{password\_strength} is greater than 0.7.

For this task, we use exactly the same combinatorial CNN as in the previous chapter. There is no need to adjust layer parameters as the words we train on are smaller than in the previous example. Again, as in the previous chapter, we do not search for the most optimal solution. Instead, we stop at the first one that works well. Our training parameters are exactly the same as for the palindrome dataset (batch = 32, 30 steps per epoch). The accuracy achieved on the training dataset after 19 epochs is $100\%$, while the accuracy on the validation dataset is $99.75\%$. 

\section{Discussion}
Having demonstrated the feasibility of using combinatorial convolutional networks for word classification, we would now like to discuss some possible limitations of our work, as well as future improvements and research directions.

Our main limitation lies in our focus on one-dimensional objects, specifically short-length words. This simplifies our work in two ways. Firstly, we have a clear combinatorial description of words, outlined in \cite{morita} that we can rely on. Secondly, we can perform computational experiments without the need for computational optimization of the code. In fact, data generation, combinatorial calculation, and neural network training can all be completed in a single day using a machine without a GPU. There is no reason to expect that the combinatorial CNN will fail in its task upon an increase in word length, as long as its layer sizes are appropriately adjusted. 

It is worth pointing out that if the task of classification is based solely on combinatorial patterns, such as in the case of palindrome detection, conventional neural networks will learn to do exactly that very efficiently. However, if providing the best results relies on combining combinatorial patterns with other features that depend on particular choices of letters or pixels, conventional neural networks may prefer such a mix, and the value of our approach will result in ablution studies, e.g. finding to what extent combinatorial patterns are important in a given situation. In cases where there is no clear preference between two options of mixed combinatorial patterns/other features and purely combinatorial patterns, one may choose our approach to build a more robust system.

Last but not least, this work demonstrates how the results of 'pure' mathematics find their way into applications. The almost immediate applicability of results stemming from the study of objects in abstract algebra and published in mathematical journals dedicated to the subject underscores the importance of such research for practical applications as well. We expect that expanding one-dimensional results to higher dimensions and investigating the role of different symmetries in deep learning may present another step forward.

\subsubsection*{Acknowledgments.} The author would like to express gratitude to Jun Morita for pointing out works on the combinatorial description of words and for providing valuable clarifications.


\begin{thebibliography}{4}

\bibitem{deepl} Ian Goodfellow, Yoshua Bengio, Aaron Courville, Deep Learning, MIT Press (2016)

\bibitem{cats} Keras "Image classification from scratch" example, \url{https://keras.io}

\bibitem{zhang} Xiang Zhang, Junbo Zhao, Yann LeCun. Character-level Convolutional Networks for Text Classification. NIPS (2015)

\bibitem{baba} Simon A. Babayan, Richard J. Orton, Daniel G. Streicker, Predicting reservoir hosts and arthropod vectors from evolutionary signatures in RNA virus genomes, Science, 362, 6414, 577 (2018)

\bibitem{iuchi} Hitoshi Iuchi, Junna Kawasaki, Kento Kubo, Tsukasa Fukunaga, Koki Hokao, Gentaro Yokoyama, Akiko Ichinose, Kanta Suga, Michiaki Hamada, Bioinformatics approaches for unveiling virus-host interactions, Computational and Structural Biotechnology Journal, Volume 21, 1774-1784 (2023)

\bibitem{key} Ortuno M , Carpena P, Bernaola-Galvan P, Munoz E, Somoza AM. Keyword detection in natural languages and DNA. Europhys. Lett. (2002) 57:759–764.

\bibitem{RNA} Karen Sargsyan, Carmay Lim, Arrangement of 3D structural motifs in ribosomal RNA, Nucleic Acids Research, Volume 38, Issue 11, 1 June 2010, Pages 3512–3522

\bibitem{morita} Jun Morita, Akira Terui, Words, tilings and combinatorial spectra, Hiroshima Math. J., 39, 37–60 (2009)

\bibitem{morita2} Jun Morita, Words, automata and Lie theory for tilings, Symmetries, Integrable Systems and Representation (eds. K. Iohara, M.-G. Sophie and B. Remy), Proc. Math. Stat. 40, Sprnger, (2013) 

\bibitem{morita3} Jun, Morita. Tilings, Lie theory and combinatorics. Contemporary Mathematics. 506. 173-185 (2010)

\end{thebibliography}
\end{document}